%% file: acl2023.tex
\pdfoutput=1
\documentclass[11pt]{article}

\usepackage[]{ACL2023}

\usepackage{times}
\usepackage{latexsym}
\usepackage{multirow}

\usepackage[T1]{fontenc}
\usepackage[utf8]{inputenc}

\usepackage{microtype}

\usepackage{inconsolata}
\usepackage{spverbatim}
\usepackage{graphicx}% http://ctan.org/pkg/graphicx
\usepackage{multirow}
\usepackage{tabularx}
\usepackage{booktabs}
\usepackage{xspace}
\usepackage{amsmath,amsfonts,amssymb}
\usepackage{fdsymbol}

\newcommand{\OurMethod}{KB-BINDER\xspace}
\newcommand{\Freebase}{{\textsc{Freebase}}\xspace}
\newcommand{\iid}{i.i.d.\xspace}

\def\ie{\textit{i.e.}\xspace}
\def\eg{\textit{e.g.}\xspace}

\newenvironment{remark}[1][Remark]{\begin{trivlist}
\item[\hskip \labelsep {\bfseries #1}]}{\end{trivlist}}

\title{Few-shot In-context Learning for Knowledge Base Question Answering}

\author{$^{\spadesuit}$Tianle Li, $^{\spadesuit}$Xueguang Ma, $^{\spadesuit}$Alex Zhuang, $^\varheartsuit$Yu Gu,  $^\varheartsuit$Yu Su, $^{\spadesuit,\clubsuit}$Wenhu Chen \\
    $^\spadesuit$University of Waterloo \\
    $^\varheartsuit$The Ohio State University \\
    $^\clubsuit$Vector Institute, Toronto \\
  \texttt{\{t29li,x93ma,a5zhuang,wenhuchen\}@uwaterloo.ca, \{gu.826,su.809\}@osu.edu} \\}

\begin{document}
\maketitle
\begin{abstract}
Question answering over knowledge bases is considered a difficult problem due to the challenge of generalizing to a wide variety of possible natural language questions. Additionally, the heterogeneity of knowledge base schema items between different knowledge bases often necessitates specialized training for different knowledge base question-answering (KBQA) datasets. To handle questions over diverse KBQA datasets with a unified training-free framework, we propose \OurMethod, which for the first time enables few-shot in-context learning over KBQA tasks. Firstly, \OurMethod leverages large language models like Codex to generate logical forms as the draft for a specific question by imitating a few demonstrations. Secondly,  \OurMethod grounds on the knowledge base to bind the generated draft to an executable one with BM25 score matching. The experimental results on four public heterogeneous KBQA datasets show that KB-BINDER can achieve a strong performance with only a few in-context demonstrations. Especially on GraphQA and 3-hop MetaQA, \OurMethod can even outperform the state-of-the-art trained models. On GrailQA and WebQSP, our model is also on par with other fully-trained models. We believe \OurMethod can serve as an important baseline for future research. Our code is available at \url{https://github.com/ltl3A87/KB-BINDER}
\end{abstract}
%Question answering over knowledge bases is conceived as a challenging problem with respect to its generalization to unseen questions. Due to the heterogeneity of the knowledge base schema items from one knowledge base to the other, it entails specialized training for different knowledge-based question-answering (KBQA) datasets. To handle questions over diverse KBQA datasets with a unified training-free framework, we propose \OurMethod, which for the first time enables few-shot in-context learning over KBQA tasks. Firstly, \OurMethod leverages large language models like Codex to generate logical forms as the draft for a specific question by imitating a few demonstrations. Secondly,  \OurMethod grounds on the knowledge base to bind the generated draft to an executable one with BM25 score matching. The experimental results on four public heterogeneous KBQA datasets show that KB-BINDER can achieve strong performance with only a few in-context demonstrations. Especially on GraphQA and 3-hop MetaQA, \OurMethod can even outperform the state-of-the-art trained models. On GrailQA and WebQSP, our model is also on par with other fully-trained models. We believe \OurMethod can serve as an important baseline for future research.

\input{introduction.tex}
\input{related.tex}
\input{methodology.tex}
\input{experiments.tex}

\input{conclusion.tex}
\input{limitations.tex}
\input{ethics.tex}

\bibliography{custom}
\bibliographystyle{acl_natbib}

\clearpage
\appendix
\section{Appendix}
\label{sec:appendix}

\subsection{Error Analysis}
\label{sec:error_ana}
To explore our method performance in each step of the pipeline, we conduct independent error analysis on the 500 subset randomly sampled from the dev set of GrailQA. Tested on \OurMethod(1) with the same setting with Section \ref{sec:implement detail}, the recalls of the entity binder and relation binder are 0.9 and 0.78 respectively. And the recall of the logic frame is 0.66 for the top 1 draft, which account for most error cases.

\end{document}

%% file: introduction.tex
\section{Introduction}
\begin{figure}[t]
\centering
\small
\includegraphics[width=0.7\linewidth]
{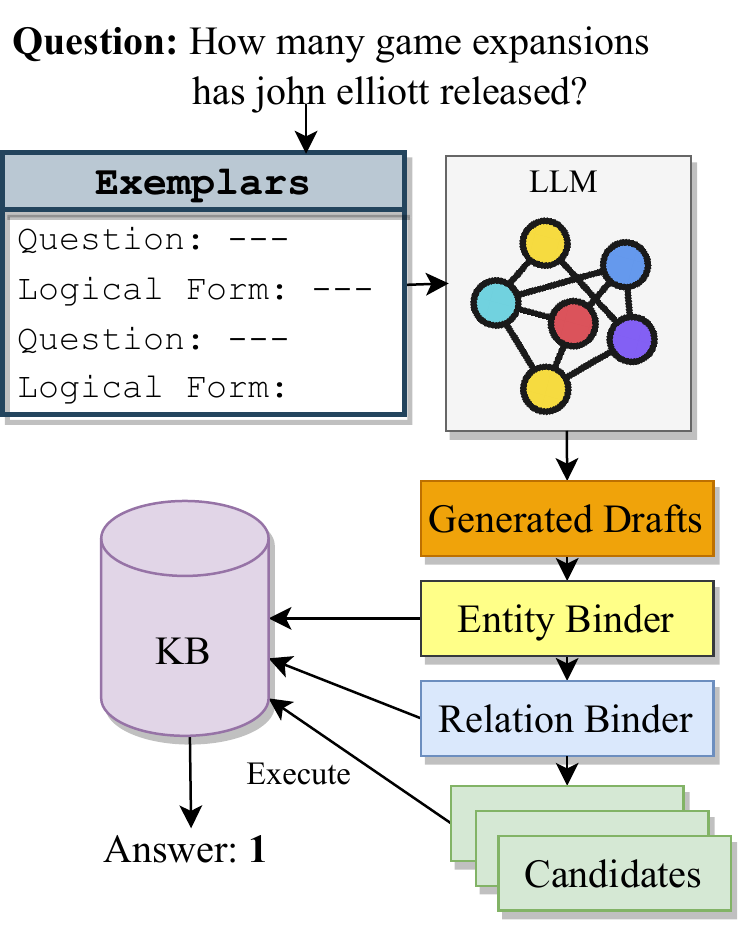}
  \caption{Overview of \OurMethod pipeline. There are two primary stages in our method: 1) Generate the drafts as preliminary logical forms; 2) Bind the drafts to the executable ones with entity and relation binders grounded on the knowledge base. The final answer can be obtained after the execution of the final candidates.}
\label{fig:model-demo}
\end{figure}
Question answering over knowledge bases (KBQA)~\cite{berant2013semantic,yih2015semantic} has been a long-standing research problem in the AI community. It has attracted wide attention from the community with its significant role in making large-scale knowledge bases accessible to non-expert users \cite{Wu2019ASO, Lan2021ASO, gu2022knowledge}. % From the aspect of users, it is an ideal scenario that they can retrieve any specific information trackable from the knowledge bases with questions in the format of natural language. 
However, despite the fact that the increasing scale of knowledge bases can enable the retrieval with higher coverage on miscellaneous topics, it poses a great challenge for suppliers with limited resources, who rely on the models trained on certain knowledge bases or benchmarks. 
Concretely, the difficulties primarily lie in the following aspects: 
1) Data intensiveness: larger knowledge bases require ever larger quantities of annotated data to allow fine-tuned models to generalize well over them. \cite{yih2016value, Talmor2018TheWA, Gu2020BeyondIT}.
2) Dataset specificity: For relatively small-scale KBQA datasets, the fully-trained models tend to overfit to a specific schema, and can hardly generalize to knowledge base questions in unseen domains \cite{Su2016OnGC, Zhang2017VariationalRF, Sun2019PullNetOD}.
%These challenges make few-shot learning crucial in KBQA, as it allows for generalization by learning from a small amount of annotated data.
%For relatively small-scale and domain-specific KBQA datasets, the trained models cannot generalize well to out-of-domain questions \cite{Su2016OnGC, Zhang2017VariationalRF, Sun2019PullNetOD}. These challenges make few-shot learning crucial in KBQA, as it allows for generalization by learning from a small amount of annotated data.
These challenges make it crucial to devise a new framework that can work in both low-resource and training-free settings in KBQA.  % learning crucial in KBQA, 

% LLM in-context is not only few-shot but also training-free.
Recently, large language models (LLMs) like GPT-3 and Codex \cite{Brown2020LanguageMA, Chen2021EvaluatingLL} have demonstrated their strong generalizability \cite{Wang2022Code4StructCG, Wei2022ChainOT, Zhou2022TeachingAR, Cheng2022BindingLM,Zhou2022LeasttoMostPE, Suzgun2022ChallengingBT} on a wide range of text, table, commonsense and even math QA tasks with few-shot in-context learning. Other works also validate that Codex~\cite{Chen2021EvaluatingLL} can parse and transform unstructured instructions to structured and executable code with only a few dozen demonstrations \cite{Gao2022PALPL, Chen2022ProgramOT}. These works inspire us to tackle KBQA with LLMs, an under-explored area in the literature that is particularly challenging compared to other QA tasks because of the massive scale of modern KBs. 

However, it is still unclear how to address KBQA with in-context learning. Unlike many other question-answering tasks, where the evidence is provided with a reasonable length limit, KBQA needs to condition on a massive graph containing millions of nodes and billions of edges. Evidently, it is impossible to feed the whole graph as-is to the language model. Even feeding a subgraph is extremely challenging as it requires splitting the monolithic graph into self-consistent and query-relevant chunks, which is itself an unaddressed research problem. Without feeding the knowledge graph as an additional input, language models become unaware of the schema of the KB. This problem makes it difficult to associate surface forms in the questions with the corresponding entities and relation types in a specific KB, not to mention generate executable logical forms with these linked entities and relations. These challenges make it hard to build in-context KBQA systems.   

% change the entity example to a real FreeBase one
In this work, we propose \OurMethod, which, for the first time, enables training-free few-shot in-context learning on KBQA.
Our framework consists of two stages as shown in Figure \ref{fig:model-demo}. 
In the first stage, we demonstrate a few KBQA questions and their corresponding logical forms as the exemplary pairs for Codex to generate a \textbf{draft} of an unseen question. The \textbf{draft} is a `preliminary' logical form likely to contain mistakes in both entities and relations. For example, due to a lack of information about the KB schema, Codex might generate a \textbf{draft} containing `medicine.manufactured\_drug.shape' while the true relation in the KB should be `medicine.manufactured\_drug\_form.shape'.
In the second stage, \OurMethod binds the `preliminary' entities to the true entity by using a lexicon-based similarity search over the whole KB. Once the entities are bound, we search through the vicinity of the bound entities to bind the `preliminary' relations. We fill the bound entities and relations into the \textbf{draft} to generate a set of `refined' logical forms. We execute these logical forms against the KB and return the executed results as the answer. To enhance \OurMethod with more pertinent exemplars, we also propose a \OurMethod-R with retrieved exemplars from the training set.

In general, previous works rely heavily on pre-defined heuristics for a target knowledge base to find the potential candidates \cite{Ye2021RNGKBQAGA, Gu2022ArcaneQADP, Shu2022TIARAMR}. \OurMethod, however, does not need heuristics customized to specific KB schema due to the inherent generalizability of LLMs. We test the performance of our models under few-shot setting on four public datasets, WebQSP \cite{yih2016value}, GrailQA \cite{Gu2020BeyondIT}, GraphQA \cite{Su2016OnGC} and MetaQA \cite{Zhang2017VariationalRF}. On GraphQA and 3-hop MetaQA, \OurMethod achieves 39.5 F1 and 99.5\% Hits@1 scores respectively, surpassing the previous SoTA by 7.7 on F1 score and 0.6\% on Hits@1 correspondingly. On WebQSP, \OurMethod-R can achieve 74.4\% F1 score, only 4.4\% lower than the SoTA model~\cite{Yu2022DecAFJD}. These experimental results demonstrate the effectiveness of our approach.

Given the simplicity and generality of \OurMethod, we believe it could serve as an important baseline for future KB research, especially in the low-resource setting.  

%% file: related.tex
\section{Related Work}

\begin{remark}[Knoweldge Base Question Answering.]
Most state-of-the-art KBQA models are based on semantic parsing~\cite{Lan2021ASO,gu2022knowledge}, where a question is mapped onto a logical form over the KB. 
Locating the target logical form over the KB entails a massive search space (\eg, \Freebase~\cite{bollacker2008freebase} contains 45 million entities and 3 billion facts).
% Early methods rely on collecting a large set of training data and test on questions following the same distribution as training (\ie, \iid assumption)~\cite{yih-etal-2015-semantic,dong-lapata-2016-language,talmor-berant-2018-web}, however, these methods are data-inefficient due to their failures in answering questions involving unseen relations during training. 
Recent methods capitalize on the strong generalizability of LMs to generalize to the massive space unexplored during training~\cite{Chen2021ReTraCkAF,Gu2022ArcaneQADP,Ye2021RNGKBQAGA,Shu2022TIARAMR}. 
These methods are more data-efficient and can better handle the massive search space compared with earlier methods operating with an \iid assumption~\cite{yih2015semantic,dong2016language}, however, they still require thousands of labeled examples to fine-tune LMs.
Despite being an appealing idea, few-shot KBQA has not been touched by existing work. It has been deemed highly non-trivial, if not impossible, to learn to handle the large search space in KBQA only with a handful of training data. The most relevant work is~\citet{Hua2020FewshotCK}, which trains a meta-model to quickly adapt to a new question with a few training examples. However, they need 2,000 labeled questions to train the meta-model first, thus not a true few-shot setting. In this paper, we present the first effort to enable true few-shot learning for KBQA with LLMs, which may point to interesting opportunities for practical KBQA under low-data settings.

\end{remark}

% \begin{remark}[KBQA.]
% There are widely known to be two categories of approaches to KBQA tasks. In information retrieval (IR) based approaches, a knowledge graph is usually extracted from the KB based on the entities detected in a question. Then, various approaches for performing reasoning over the graph are used to arrive at an answer. In semantic parsing (SP) based approaches, A question is semantically parsed into a logical form that can be grounded to a query that is directly executable on a KB, providing the answer to the question. We direct the reader to (the survey paper) for a more in-depth treatment survey of different works within each category. Our work follows the latter SP approach and is related to recent work on methods in neural program induction (NPI) in which natural language is transformed into executable programs using neural models. Discussion of () () (). Each of these methods uses a pretraining step to learn general knowledge about the datasets before few shot examples can be applied in an inference setting. One benefit of our approach is that we do not need to pretrain on any dataset. Our multi-hop approach is common to many of the grounding steps used in these other papers.
% \end{remark} 

\begin{remark}[In-Context Learning with LLMs.]
In-context learning with large language models~\cite{Brown2020LanguageMA} has shown strong few-shot performance in many NLP tasks, such as question answering \cite{Cheng2022BindingLM}, information extraction \cite{alex2022structured}, and numerical reasoning \cite{lewkowycz2022solving}. Analyses into the mechanisms behind this behavior are undertaken by \citet{olsson2022context, xie2021incontext}. Empirically, \citet{min2022rethinking} shows the effectiveness of constructing prompts using an input-label pairing format, and \citet{liu2021InContextExamples} experiment with the number of examples provided, as well the idea of retrieving relevant examples to a test input to construct the prompt with. These results inform the prompt-construction methods used in our work. \citet{Lampinen2022explanations} suggests that incorporating explanatory task instructions in context can improve performance, however, we leave a deeper exploration of this to future works.

\end{remark}

\begin{remark}[Reasoning with LLMs.]

A number of methods have recently emerged to extend the reasoning capabilities of LLMs ~\cite{Brown2020LanguageMA, kojima2022large}. Chain of Thought Prompting (CoT) \cite{Wei2022CoT} showed that encouraging intermediate steps in model output can improve reasoning accuracy. Developing this idea, methods that involve a direct synthesis of formal programs that solve these tasks have shown further improvement ~\cite{Chen2022ProgramOT, nye2021work, Gao2022PALPL, Cheng2022BindingLM}.
%Most current works focus on solving tasks where the context about operations required (i.e. performing arithmetic) can be roughly summarized in an input prompt, however, KBQA is a task where the breadth of information in the KB may be required for test examples, but cannot be provided to the LLM in a limited input space.
The most relevant work to the QA setting is Binder~\cite{Cheng2022BindingLM}, where the LLM is prompted to conduct text-to-SQL generation and further answer questions using information retrieved from an SQL database.
However, while SQL table headers demonstrated in examples can help an LLM generate reasonable SQL commands, the thousands of relations and millions of entities in a KB represent a much larger search space that cannot be captured as easily by the prompting an LLM. KB-BINDER solves this challenge using a draft generation and schema binding pipeline.
\end{remark}

% In-context learning (ICL) has been shown to be effective in enabling LLMs to perform well on various downstream tasks. By conditioning, the model at inference time with a few demonstrations in the prompt or input, (these people) have shown that this approach can be comparable to fine-tuning models directly. (These people) showed that good in-context samples usually involve these things. (These people) hypothesize that it works because of this. In this work, we compose prompts that take advantage of ICL to generate logical forms corresponding to the semantic meaning of natural language questions.  
% \end{remark}

% \begin{remark}[Reasoning with LLMs.]
% LLMs have been shown to perform well on tasks that would require human reasoning. \citeauthor{}COT did something. scratchpad did something. Binder did something. POT proposed something as well. Various other people have discussed the mechanisms for what constitutes this reason. In our work, we seek to answer whether these latent capabilities can drastically simplify the semantic parsing step in grounding KB-executable queries. To the best of our knowledge, we don't know of any prior approaches that leverage these reasoning capabilities to solve the semantic parsing problem in a strictly few-shot setting without pretraining. The non-trivial task of generating a logical form for a given natural language query that can be easily grounded to a KB query implicitly involves 
% \end{remark}

%% file: methodology.tex
\begin{figure*}[htbp]
%\centering
\includegraphics[width=\textwidth]
{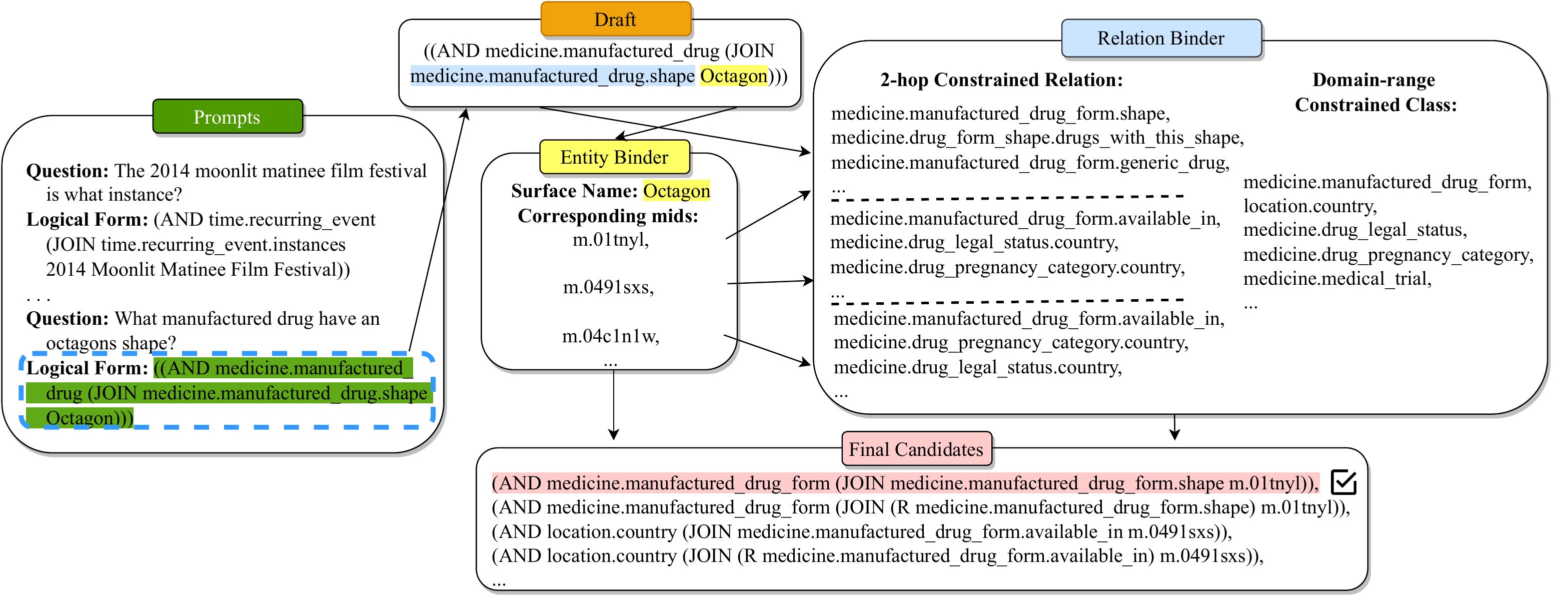}
  \caption{\OurMethod framework: Given a question, the LLM will first generate its corresponding preliminary logical forms as the drafts, imitating the exemplary demonstration. Then the entity and relation binders will operate on the drafts to ground the entities and relations on KB respectively, which produces the final candidates.}
\label{fig:draft-binder}
\end{figure*}

\section{Methodology}
Given a new question, \OurMethod leverages an LLM to generate a preliminary logical form as a draft.
A draft is not guaranteed to be executable, as it is generated by the LLM without being explicitly restricted to the candidates' vocabulary and knowledge graph structure. However, with the demonstration of in-context prompting, drafts can reveal the structural relationships among mentioned entities in a semantically reasonable way. As a result, the generated drafts can simplify the search space needed to retrieve real entities and schema terms. These entities and terms are then used to revise the draft to a real executable logical form for a given question. This process is illustrated in Figure \ref{fig:draft-binder}.

\subsection{Drafts Generator}

We leverage the in-context learning capability of Codex to generate logical form drafts for unseen questions. Specifically, we randomly sample $N$ examples from the training set as the exemplars, which are shown to the LLM in the form of <Qustion, Logical form> pairs. However, it is worth noting that the \textbf{MIDs} (i.e., machine identifier) in the original logical form are not easy to interpret and imitate. For instance, the raw logical form of the question ``data compression is the genre of which file format?" is:

\begin{small}
\begin{verbatim}
(AND computer.file_format (JOIN 
computer.file_format.genre m.0279m))
\end{verbatim}
\end{small}

% \textit{(AND computer.file\_format (JOIN computer.file\_format.genre m.0279m))}
\noindent where \textit{m.0279m} is the MID of the entity ``data compression" from FreeBase. The raw format of MIDs with no semantic meaning can hardly assist the large language model to understand and imply the latent relationships among schema items. Therefore, naturally, we substitute the MIDs in the original logical forms with their surface names in the prompting demonstrations. Consequently, the final processed logical form fed to Codex for the above example will become:

\begin{small}
\begin{verbatim}
(AND computer.file_format (JOIN 
computer.file_format.genre Data Compression))
\end{verbatim}
\end{small}

\noindent the surface names of the entities mentioned in a new target question will appear in the generated preliminary logical forms as shown in the demonstration. Through in-context learning, LLM is tasked with generating such friendly logical forms for a new question by following the demonstrations. 

\subsection{Knowledge Base Binder}
\label{sec:LF_binder}
The preliminary logical forms generated by the large language model provide us with a macroscopic view of the question from the perspective of semantics and structure relationships. Starting from the generated drafts, we separately perform the entity and relation binding over the KB.
\paragraph{Entity Binder}
To identify the exact MIDs of the entities mentioned in the questions, we directly extract their surface names from the generated drafts. If the extracted surface names consistently match the friendly names of some MIDs from the knowledge base, we retrieve all the MIDs corresponding to the matched friendly names and select the most popular
$n$ of them based on FACC1. If the surface names match no friendly name of any entity from the knowledge base, we then utilize BM25 to retrieve the most similar existing one in KB and exploit it as the anchor to extract the MID candidates.
% first. If there are fewer such MIDs than a threshold $n$, we employ all these MIDs as the potential candidates for the target entity. Otherwise, we retrieve all the possible types for each of these MIDs, and use the question as the search query to retrieve the most related type compared to the question. Subsequently, we capture the top $n$ entities within the most related type as the final candidates. If the surface names match no friendly name of any entity from the knowledge base, we then utilize BM25 to retrieve the most similar existing friendly name in the knowledge base and exploit it as the anchor to retrieve the most possible mid candidates with similar steps in the first case.
If we detect multiple surface names from the drafts, we bind their potential MIDs independently first. And all the permutations of their combinations will be considered in the final execution.

\paragraph{Relation Binder}
In spite of the fact that the generated preliminary relations in the drafts are very likely to not exist in the knowledge base, their format and semantic meaning are still supposed to be analogical to the real-existed ones, learning from the demonstration of the prompts. With this assumption, we utilize each of the related items together with the original question as the search query to retrieve the most similar ones with BM25 from the whole knowledge base relation collection. To enlarge the possibility of successful execution of the logical form, we only keep the top $m$ among all the two-hop relation items starting from the MIDs of the current permutation and filter out the ones out of this constraint. For each combination of MIDs, we iterate all the $m$ retrieved relations candidates accordingly.

% \subsubsection{Frame Binder}
% After obtaining the entity and relation candidates, the last step is to verify whether the operation flows in the drafts are correct. We simply traverse all the combinations of reverse operation \textbf{R} and join operation \textbf{JOIN} that appeared in the generated drafts. Specifically, for a draft of one \textbf{JOIN} operation, the final candidates will contain a version with and without \textbf{R} operation exerted on the relation item after \textbf{JOIN}.
\paragraph{Majority Vote}
Following the above workflow, a generated draft can be bound to hundreds of potential logical form candidates. And each of them can be converted to a SPARQL query to be ultimately executed on the KB. We record all the answerable logical form candidates and their corresponding answers. As self-consistency can improve the robustness of the predictions of large language model \cite{Wang2022SelfConsistencyIC}, we repeat the paradigm for $K$ times and adapt the majority vote strategy to decide the final consistent answer and its logical form. We name the model with self-consistency on the top $K$ drafts as KB-BINDER(K).

\paragraph{Retrieved Exemplars}
\label{sec:retrieval-augmented}
 To further boost the performance of our method in a training-free setting, we design another variant of KB-BINDER, named KB-BINDER(K)-R. Instead of selecting the exemplars from the training sets randomly, KB-BINDER(K)-R leverages BM25 to retrieve the most similar $N$ questions with the target one as the demonstrations. So that the logical forms of the $N$ questions are more likely to cover the schema items that are related or even exactly the same as the target one. This setting is supposed to be especially advantageous over questions of I.I.D. type.

%% file: experiments.tex
\section{Experiment}
In this section, we briefly introduce the benchmarks used to evaluate the performance of our framework. And we demonstrate the detailed setting of \OurMethod and its result on each of the datasets compared with the fully-trained baselines. Ultimately, we make an analysis of the variation of design choices and their corresponding potential causes.
\subsection{Datasets}

\begin{table}[t]
\small
\centering
\begin{tabular}{l|lll}
\toprule
\textbf{Dataset} & \textbf{Train} & \textbf{Dev} & \textbf{Test} \\
\midrule
GrialQA & 44,337 & 6,763 & 13,231 \\
WebQSP  & 3,098 & $-$ & 1,639 \\
GraphQA & 2,381  & $-$ & 2,395 \\
MetaQA-1hop  & 96,106 & 9,992 & 9,947 \\
MetaQA-2hop  & 118,980 & 14,872 & 14,872 \\
MetaQA-3hop  & 114,196 & 14,274 & 14,274 \\
\bottomrule
\end{tabular}
\caption{Dataset statistics.}
\label{table:dataset}
\end{table}

We evaluate \OurMethod on four public KBQA datasets as follows: 

\noindent \textbf{GrailQA}~\cite{Gu2020BeyondIT} is a diverse KBQA dataset built on Freebase, covering 32,585 entities, 3,720 relations across 86 domains. It is designed to test three levels of generalization of KBQA models: I.I.D., compositional, and zero-shot.
% It covers 32,585 entities and 3720 relations under 86 domains, and at most four relations can be embedded in one question.
% Strongly Generalizable Question Answering (GrailQA) is a highly diverse KBQA datasets based on Freebase. It covers 32,585 entities and 3720 relations under 86 domains, and at most four relations can be embedded in one question. 
% Strongly Generalizable Question Answering (GrailQA) is a new large-scale, high-quality dataset for question answering on knowledge bases (KBQA) on Freebase, annotated with SPARQL, S-expression. It can be used to test three levels of generalization in KBQA: i.i.d., compositional, and zero-shot.
% Following prior work, we generate logical forms containing up to 4 relations and optionally containing one function selected from counting, superlatives (argmax, argmin), and comparatives (>, ≥, <, ≤). 

\noindent \textbf{GraphQA}~\cite{Su2016OnGC} is also a diverse dataset that covers a wide range of domains. It builds by sentence-level paraphrasing from graph queries and evaluating compositional generalization.
% GRAPHQUESTIONS contains 500 graph queries, 2,460 sentence-level paraphrases, and 5,166 questions2. The dataset presents a high diversity and covers a wide range of domains including People, Astronomy, Medicine, etc. Specifically, it contains 148, 506, 596, 376 and 3,026 distinct domains, classes, relations, topic entities, and words, respectively. We evenly split GRAPHQUESTIONS into a training set and a testing set. All the paraphrases of the same graph query are in the same set

\noindent \textbf{WebQSP}~\cite{yih2016value} contains questions from WebQuestions that are answerable by Freebase.
It tests i.i.d. generalization on simple questions.

\noindent \textbf{MetaQA}~\cite{Zhang2017VariationalRF} consists of a movie ontology derived from the WikiMovies Dataset and three sets of question-answer pairs written in different levels of difficulty. It evaluates the effectiveness in a specific domain.

\noindent Table~\ref{table:dataset} shows the detail of train/dev/test splits of the datasets.
We evaluate our pipeline on all the test sets and conduct ablation studies on a subset of the dev set from GrailQA with 500 randomly sampled examples.

\subsection{Baselines}
We compare our method with all the systems that have a publication on the official leaderboard of each dataset and record their results from the paper directly with the same evaluation matrix.
Notice that all the competitive baseline methods utilized the entire set of training data as supervision.

\subsection{Implementation Details}
\label{sec:implement detail}
In the draft generation step, we leverage \texttt{code-davinci-002} from OpenAI API\footnote{https://openai.com/blog/openai-codex/} to obtain the top $K$ drafts for each question, we test the cases with $K=1$ and $K=6$, and refer to them as \OurMethod(1) and \OurMethod(6). Specifically, we randomly sample $N=100$ exemplary questions from the training sets of WebQSP and GraphQA respectively. For GrailQA, we sample $N=40$ exemplars for testing due to the long inference time on more than ten thousands of testing data. For MetaQA, we only sample 5 questions for demonstration, as the KB is relatively small in this benchmark.
% we randomly sample 40 questions from the training sets of GrailQA, WebQSP and GraphQA as the demonstrations in in-context learning for GrailQA, WebQSP and GraphQA. For MetaQA, we randomly choose 20, 5 and 5 questions from 1-hop, 2-hop and 3-hop Vanilla training sets respectively, as all the questions from MetaQA is under the same domain. 

In the binding step, we set $n=15$ for all the questions in the entity binder. We deploy BM25 and Contriever \cite{Izacard2021UnsupervisedDI} provided by Pyserini\footnote{https://github.com/castorini/pyserini} as a hybrid searcher to retrieve the originally unmatched friendly names and the top relation items. After obtaining the globally ranked relations, we focus on the relations bound by 2-hop relations from the detected entities. We traverse the top 10 (i.e., $m=10$) relation candidates within the 2-hop constraint for GrailQA, WebQSP and GraphQA, and the top 1 (i.e., $m=1$) for MetaQA. After the drafts are bound to the potential candidates, they will be translated to SPARQL and executed on the Virtuoso server following the instructions\footnote{https://github.com/dki-lab/Freebase-Setup}.

\begin{table}[t]
\small
\centering
\begin{tabular}{lcc}
\toprule
&\multicolumn{2}{c}{\textbf{Overall}} \\
 \cmidrule(lr){2-3}
\multicolumn{1}{l}{\textbf{Method}} & \textbf{EM} & \textbf{F1} \\ 
\midrule
\multicolumn{1}{l}{GloVe + Transduction~\citep{Gu2020BeyondIT}} & 17.6  & 18.4 \\
\multicolumn{1}{l}{QGG~\citep{Lan2020QueryGG}} & - & 36.7 \\
\multicolumn{1}{l}{BERT + Transduction~\citep{Gu2020BeyondIT}} & 33.3  & 36.8 \\
\multicolumn{1}{l}{GloVe + Ranking~\citep{Gu2020BeyondIT}}     & 39.5  & 45.1 \\
\multicolumn{1}{l}{BERT + Ranking~\citep{Gu2020BeyondIT}}      & 50.6  & 58.0 \\
\multicolumn{1}{l}{ReTraCk~\citep{Chen2021ReTraCkAF}} & 58.1 & 65.3  \\
\multicolumn{1}{l}{S$^{2}$QL~\citep{Zan2022S2QLRA}}   & 57.5 & 66.2 \\
\multicolumn{1}{l}{ArcaneQA~\citep{Gu2022ArcaneQADP}}  & 63.8 & 73.7 \\
\multicolumn{1}{l}{RnG-KBQA~\citep{Ye2021RNGKBQAGA}}  & 68.8 & 74.4 \\
\multicolumn{1}{l}{DecAF~\citep{Yu2022DecAFJD}} & 68.4 & 78.7 \\
\multicolumn{1}{l}{TIARA~\citep{Shu2022TIARAMR}} & \textbf{73.0} & \textbf{78.5} \\
\midrule
\multicolumn{1}{l}{\textbf{Few-shot in-context}} \\
\multicolumn{1}{l}{~~~~~~\textbf{\OurMethod(1)}} & 47.0 & 51.6 \\
\multicolumn{1}{l}{~~~~~~\textbf{\OurMethod(6)}} & 50.6 & 56.0 \\
\multicolumn{1}{l}{~~~~~~\textbf{\OurMethod(6)-R}} & 53.2 & 58.5 \\
\bottomrule
\end{tabular}
\caption{40-shot Results of \OurMethod/\OurMethod-R and baselines on GrailQA.}
\label{table:grailqa test results}
\end{table}

\begin{table}[ht]
% \footnotesize
\centering
\small
\begin{tabular}{lc}
\toprule
\multicolumn{1}{l}{\textbf{Method}} & \textbf{F1} \\ 
\midrule
\multicolumn{1}{l}{ReTraCk~\citep{Chen2021ReTraCkAF}} & 71.0 \\
\multicolumn{1}{l}{QGG~\citep{Lan2020QueryGG}} & 74.0 \\
\multicolumn{1}{l}{ArcaneQA~\citep{Gu2022ArcaneQADP}} & 75.6 \\
\multicolumn{1}{l}{PullNet~\citep{Sun2019PullNetOD}} & 62.8 \\
\multicolumn{1}{l}{RnG-KBQA~\citep{Ye2021RNGKBQAGA}} & 75.6 \\
\multicolumn{1}{l}{TIARA~\citep{Shu2022TIARAMR}} & 76.7 \\
\multicolumn{1}{l}{DecAF~\citep{Yu2022DecAFJD}} & \textbf{78.8} \\
\midrule
\multicolumn{1}{l}{\textbf{Few-shot in-context}} \\
\multicolumn{1}{l}{~~~~~~\textbf{\OurMethod(1)}} & 52.5 \\
\multicolumn{1}{l}{~~~~~~\textbf{\OurMethod(6)}} & 53.2 \\
\multicolumn{1}{l}{~~~~~~\textbf{\OurMethod(6)-R}} & 74.4 \\
\bottomrule
 \end{tabular}
    \caption{100-shot Results of \OurMethod/\OurMethod-R and baselines on WebQSP.}
    \label{table:webqsp_main}
\end{table}

\begin{table}[ht]
\centering
% \footnotesize
\small
\begin{tabular}{lc}
\toprule
\multicolumn{1}{l}{\textbf{Method}} & \textbf{F1} \\ 
\midrule
\multicolumn{1}{l}{AUDEPLAMBDA~\citep{Reddy2017UniversalSP}} & 17.7 \\
\multicolumn{1}{l}{SPARQA~\citep{Sun2020SPARQASS}} & 21.5 \\
\multicolumn{1}{l}{BERT + Ranking~\citep{Gu2020BeyondIT}} & 25.0 \\
\multicolumn{1}{l}{ArcaneQA~\citep{Gu2022ArcaneQADP}} & 31.8 \\

\midrule
\multicolumn{1}{l}{\textbf{Few-shot in-context}} \\
\multicolumn{1}{l}{~~~~~~\textbf{\OurMethod(1)}} & 39.3 \\
\multicolumn{1}{l}{~~~~~~\textbf{\OurMethod(6)}} & \textbf{39.5} \\
\multicolumn{1}{l}{~~~~~~\textbf{\OurMethod(6)-R}} & 38.7 \\
\bottomrule
 \end{tabular}
    \caption{100-shot Results of \OurMethod/\OurMethod-R and baselines on GraphQA.}
    \label{table:graphqa_main}
\end{table}

\begin{table}[ht]
\centering
% \footnotesize
\small
\begin{tabular}{lccc}
\toprule
\multicolumn{1}{l}{\textbf{Method}} & \textbf{1-hop} & \textbf{2-hop} & \textbf{3-hop} \\ 
\midrule
\multicolumn{1}{l}{KV-Mem~\citep{Miller2016KeyValueMN}} & 96.2 & 82.7 & 48.9\\
\multicolumn{1}{l}{VRN~\citep{Zhang2017VariationalRF}} & \textbf{97.5} & 89.9 & 62.5\\
\multicolumn{1}{l}{GraftNet~\citep{Sun2018OpenDQ}} & 97.0 & 94.8 & 77.7\\
\multicolumn{1}{l}{PullNet~\citep{Sun2019PullNetOD}} & 97.0 & \textbf{99.9} & 91.4\\
\multicolumn{1}{l}{Emb~\citep{Saxena2020ImprovingMQ}} & \textbf{97.5} & 98.8 & 94.8\\
\multicolumn{1}{l}{NSM~\citep{He2021ImprovingMK}} & 97.1 & \textbf{99.9} & 98.9\\

\midrule
\multicolumn{1}{l}{\textbf{Few-shot in-context}} \\
\multicolumn{1}{l}{~~~~~~\textbf{\OurMethod(1)}} & 93.5 & 99.6 & 96.4 \\
\multicolumn{1}{l}{~~~~~~\textbf{\OurMethod(1)-R}} & 92.9 & \textbf{99.9} & \textbf{99.5} \\
\bottomrule
 \end{tabular}
    \caption{5-shot Results of \OurMethod/\OurMethod-R and baselines on MetaQA.}
    \label{table:metaqa_main}
\end{table}

\subsection{Main Result}
We demonstrate the model performance on the test sets of four public datasets in Table \ref{table:grailqa test results}, \ref{table:webqsp_main}, \ref{table:graphqa_main} and \ref{table:metaqa_main} for GrailQA, WebQSP, GraphQA and MetaQA respectively. \OurMethod(1) refers to our method in default-setting with top 1 draft, and \OurMethod(6) involves mass voting to achieve self-consistency with top 6 drafts, while \OurMethod(6)-R refers to \OurMethod(6) using retrieved exemplars \ref{sec:retrieval-augmented}. In general, all the variations of \OurMethod have strong performance on all the selected datasets. According to the results from the tables, \OurMethod(6) can generally outperform \OurMethod(1) in line with our expectations, while \OurMethod(6)-R can further boost the performance in most of the cases. And we observe that our few-shot method can achieve on par and even better performances compared to the fully supervised SOTAs on WebQSP, GraphQA and MetaQA, and it shows competitive performance with the BERT-ranking baseline on GrailQA.
% Table \ref{table:grailqa test results}
% Table \ref{table:webqsp_main}

\paragraph{\OurMethod Results} 
Specifically, we show \OurMethod(K) few-shot result on GrailQA and compare with it a series of fully-trained baselines in Table \ref{table:grailqa test results}.
With merely 40 examples, \OurMethod(6) achieves 50.6 EM score, which is the same as BERT + Ranking setting, finetuned on the whole training sets with around 45k annotations.
Although the overall scores of the two systems are on par, we notice from Table \ref{table:grailqa analysis} that our pipeline has better generalization performance on compositional and zero-shot questions, where the specific logical form is unseen in the training data.
The EM scores of \OurMethod(6) for compositional and zero-shot questions are 5.1 and 1.3 points higher than BERT+Ranking~\autoref{table:grailqa analysis}. 
We notice there is a gap between our method and the state-of-the-art supervised methods on GrailQA, however, it is still exciting to see few-shot methods is at the level of supervised methods.
%We also observe that for all the supervised baselines, there is a relatively big gap between I.I.D. typed question and the other two types (\ie{ the decreased EM scores ranging from 10 to 47.5 points}). But with our method, the performances are stable among all types.

As shown in Table~\ref{table:graphqa_main}, \OurMethod(1) and \OurMethod(6) achieve 39.3 and 39.5 F1 score on GraphQA dataset, surpassing the previous sate-of-the-art models 7.7 in F1 score. In Table~\ref{table:metaqa_main}, \OurMethod(1) achieves 99.6 \% and 96.4 \% Hits@1 scores on 2-hop and 3-hop MataQA dataset correspondingly, which are on par with the state-of-the-art models. These competitive performances show the advantage of \OurMethod on some special scenarios. For the case of GraphQA, it has a relatively small scale of training examples (\ie, 2,381 in total), however, all the questions in the test set are of compositional type. Therefore, it is hard for the fine-tuned models to become generally adapted to the novel composition of schema items, but relatively easier for LLM to generalize on this situation. For the case of MetaQA, the scale of the knowledge base (\ie, {WikiMovies}) involved in the dataset is relatively small with only dozens of unique relations under the same domain. In this case, the context and topic of the demonstration match exactly the target one, so five demonstrations are enough for LLM to generate highly accurate preliminary relation candidates.

\paragraph{\OurMethod-R Results}
As recorded in Table \ref{table:metaqa_main}, \OurMethod(1)-R sets new SoTA Hits@1 score on 3-hop MetaQA as 99.5 \%, and it achieves exactly the same performance with the previous fully-trained SoTA on 2-hop MetaQA as 99.9 \%.
From the recordings on all the tables,  we observe that \OurMethod(K)-R has a generally better performance than \OurMethod(K). Nevertheless, it is worth noting that the improvement on GrailQA is only 2.6 points, while the performance is even slightly weakened on GraphQA by 0.8 points. But \OurMethod(K)-R dramatically increases the F1 score from 53.2 to 74.4 on WebQSP. It can be rationally explained by the inherent characteristics of the datasets that GrailQA is largely composed of compositional and zero-shot questions and GraphQA only contains compositional questions, while all the questions of the test set on WebQSP are of I.I.D type, which makes the unseen questions more similar to the retrieved exemplars. 

In a nutshell, according to the presented experiment results, few-shot approaches with LLMs, such as \OurMethod(K) can at least achieve performance on par with previous fully-trained SoTAs on KBQA tasks in the following two situations: 1) There is no large-scale annotated training data, but the inference requires high generalizability of the model (i.e., GraphQA); 2) The knowledge base and the corresponding questions are very specific to one domain, so that the search space of the schema items is relatively small, but the inference requires multi-hop reasoning (i.e., MetaQA). And when it comes to a totally I.I.D setting (i.e., WebQSP), \OurMethod(K)-R can boost the performance to be on par with the supervised models. However, for the case of a large amount of training data with a high requirement for generalizability during inference (i.e., GrailQA), the previous models may have advantages over \OurMethod due to the fact that the coverage of logical form structures and schema items is restricted in our method.

\begin{figure}[t]
\centering
\includegraphics[width=0.9\linewidth]
{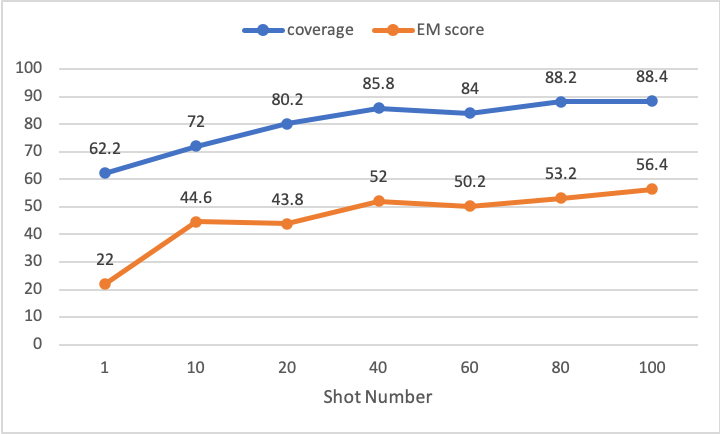}
  \caption{\OurMethod coverage and EM scores trend with shot number.}
\label{fig:shot-num}
\end{figure}

\begin{figure}[t]
\centering
\includegraphics[width=0.9\linewidth]
{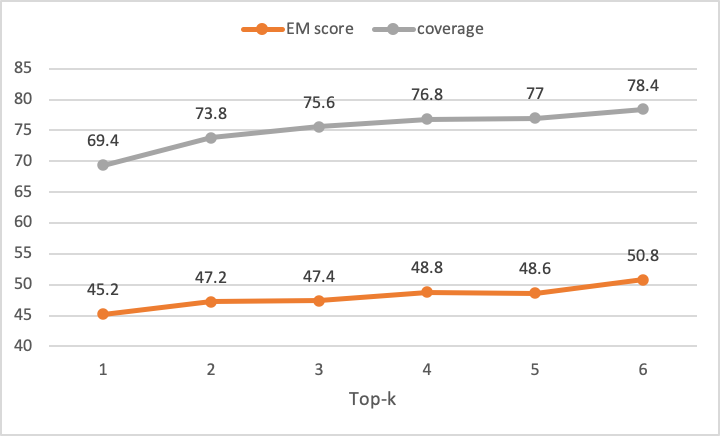}
  \caption{\OurMethod coverage and EM scores trend with top K self-consistency.}
\label{fig:topk}
\end{figure}

\subsection{Ablation Study}

\begin{table*}[t]
\small
\centering
\begin{tabular}{lcccccc}
\toprule
&\multicolumn{2}{c}{\textbf{IID}}  &\multicolumn{2}{c}{\textbf{Compositional}}    &\multicolumn{2}{c}{\textbf{Zero-shot}}            \\
 \cmidrule(lr){2-3} \cmidrule(lr){4-5} \cmidrule(lr){6-7}
\multicolumn{1}{l}{\textbf{Method}} & \textbf{EM} & \textbf{F1} & \textbf{EM} & \textbf{F1} & \textbf{EM} & \textbf{F1} \\ 
\midrule
\multicolumn{1}{l}{GloVe + Transduction~\citep{Gu2020BeyondIT}} & 50.5  & 51.6 & 16.4  & 18.5     & ~~3.0  & ~~3.1    \\
% \multicolumn{1}{l}{QGG~\citep{Lan2020QueryGG}} & - & 40.5  & -  & 33.0  & - & 36.6    \\
% \multicolumn{1}{l}{BERT + Transduction~\citep{Gu2020BeyondIT}} & 51.8  & 53.9  & 31.0  & 36.0     & 25.7  & 29.3    \\
% \multicolumn{1}{l}{GloVe + Ranking~\citep{Gu2020BeyondIT}}     & 62.2  & 67.3 & 40.0  &  47.8 & 28.9  & 33.8  \\
\multicolumn{1}{l}{BERT + Ranking~\citep{Gu2020BeyondIT}}      & 59.9  & 67.0 & 45.5  &  53.9  & 48.6  & 55.7  \\
%\multicolumn{1}{l}{ReTraCk~\citep{Chen2021ReTraCkAF}} & 84.4 & 87.5 & 61.5 & 70.9 & 44.6 & 52.5  \\
% \multicolumn{1}{l}{S$^{2}$QL~\citep{Zan2022S2QLRA}}   & 65.1 & 72.9 & 54.7 & 64.7 & 55.1 & 63.6 \\
%\multicolumn{1}{l}{ArcaneQA~\citep{Gu2022ArcaneQADP}}  & 85.6 & 88.9 & 65.8 & 75.3 & 52.9 &  66.0 \\
\multicolumn{1}{l}{RnG-KBQA~\citep{Ye2021RNGKBQAGA}}  & 86.2 & 89.0 & 63.8 & 71.2 & 63.0 & 69.2 \\
% \multicolumn{1}{l}{DecAF~\citep{Yu2022DecAFJD}} & 84.8 & 89.9 & \textbf{73.4} & \textbf{81.8} & 58.6 & 72.3 \\
\multicolumn{1}{l}{TIARA~\citep{Shu2022TIARAMR}} & \textbf{87.8} & \textbf{90.6} & 69.2 & 76.5 & \textbf{68.0} & \textbf{73.9} \\
\midrule
\multicolumn{1}{l}{\textbf{Few-shot in-context}} \\
%\multicolumn{1}{l}{~~~~~~\textbf{\OurMethod} (greedy)} & 0 & 0 & 0 & 0 & 0 & 0 & 0 & 0 \\
\multicolumn{1}{l}{~~~~~~\textbf{\OurMethod(6)}} & 51.9 & 57.4 & 50.6 & 56.6 & 49.9 & 55.1 \\
\multicolumn{1}{l}{~~~~~~\textbf{\OurMethod(6)-R}} & 72.5 & 77.4 & 51.8 & 58.3 & 45.0 & 49.9 \\
% \multicolumn{1}{l}{\textbf{TIARA} } & $72.1$ & $77.5$ & $87.2$ & $90.2$ & $68.9$ & $76.1$ & $66.6$ & $72.3$ \\
\bottomrule
\end{tabular}
\caption{Results of \OurMethod/\OurMethod-R and baselines on different question types of GrailQA.}
\label{table:grailqa analysis}
\end{table*}

\begin{figure*}[t]
\centering
\includegraphics[width=\linewidth]
{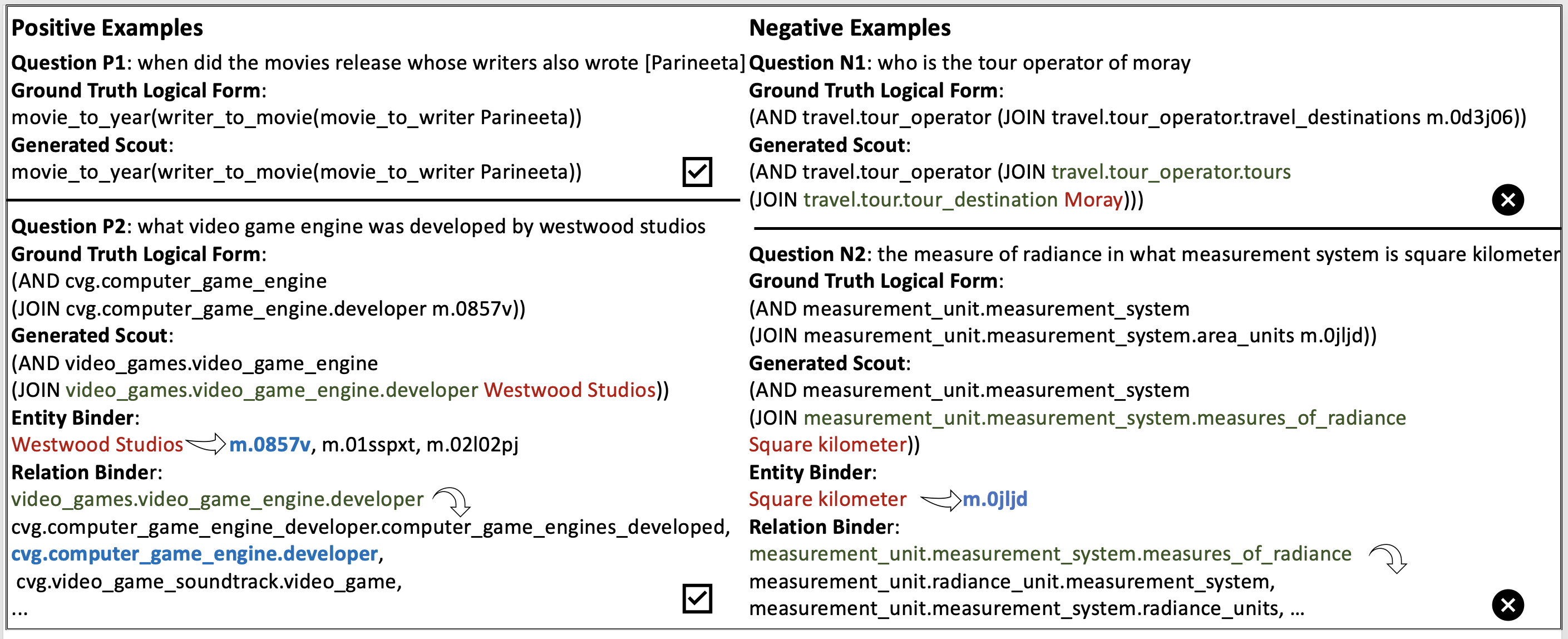}
  \caption{Positive and negative examples generated by \OurMethod.}
\label{fig:case}
\end{figure*}

We conduct ablation studies to understand the influence of the number of examples demonstrated during drafts generation on the final EM score. Due to the long inference time to complete all the testing questions, we evaluate the performance on 500 randomly sampled questions from the dev set of GrailQA. We set the number of few shot exemplars from 1 to 100, and test the coverage and EM score on each choice. The coverage here refers to the number of questions that can be grounded to at least one executable logical form over the total number of questions in the sampled set. As shown in Figure \ref{fig:shot-num}, there is an apparent trend that both the coverage and the EM score will increase with a larger number of examples. 
%However, as the number of demonstration increases, the inference speed and cost will also increase consequently.

Similarly, we also test \OurMethod(K) performance with respect to the different numbers of top drafts generated by Codex to perform the majority voting. With 40 exemplars, the result is plotted as Figure \ref{fig:topk}. Generally, increasing the number of drafts from 1 to 6 can contribute to an improvement of coverage by 19\% and EM score by 5.6\%. As if there are more drafts, more logical form structures and more formats of preliminary schema items can be covered in the first place.

However, it is also worth noting that increasing the number of shots and the number of generated drafts can also increase the inference time and cost for \OurMethod to find the answer. Taking this reason into account, we only report the results of 40 exemplars with the top 6 drafts on GrailQA, as there is always a trade-off between accuracy and the cost of time. And it also implies that there is still space for improvement for \OurMethod if we increase both of the parameters.

Moreover, we also observe from Table \ref{table:grailqa analysis} that for all the supervised baselines, there is a relatively big gap between I.I.D. typed questions and the other two types (\ie, the decreased EM score ranging from 10 to 47.5 points). 
But with \OurMethod, the performances are stable among all the types. This is due to the fact that all the questions may not come from I.I.D type for few-shot setting, so there is rarely bias among the three types.

\subsection{Case Study}
% We investigated the generation pipeline of \OurMethod,

In Figure~\ref{fig:case}, we show representative correct and error cases in the \OurMethod pipeline.
For Question P1, the generated logical form could exactly match the target one.
While for Question P2, it generates the draft in correct logic but the hallucinated entity names and relations need an extra binding step to locate the executable logical form.
Question N1 is an error case where the draft does not generate correct logic. On the other hand, Question N2 gets draft logic generated correctly but grounded into wrong entities or relations.
We conduct error analysis in each step described in \ref{sec:error_ana}
% this error can potentially be fixed by a better retrieval system.

%% file: conclusion.tex
\section{Conclusion}
\OurMethod is the first framework that enables the challenging few-shot learning on KBQA with the reasoning capability of large language models. It first generates drafts with LLM as preliminary logical forms, and then binds the entities and schema items of the drafts to the target knowledge base iteratively until an executable one can be found. \OurMethod(K) adopts majority voting, further enlarging the proportion of answerable questions with the help of more diverse formats of top K drafts. \OurMethod(K)-R with retrieved exemplars is proved to be especially advantageous when applied to I.I.D questions. In general, \OurMethod and its derivatives achieve strong performance on all the common-used KBQA datasets we select, and we hope it can set a strong baseline for future work on KBQA with a low-resource setting.

%% file: limitations.tex
\section*{Limitations}
As in-context learning with LLM heavily depends on the selected exemplars in the prompt, the performance of \OurMethod might vary from different subsets of randomly sampled examples, especially in a low-shot setting. But \OurMethod still shows strong performance on thousands of data points on each testing dataset with randomly sampled exemplars, which verifies the robustness of our method to a degree.  In the meantime, the performance of \OurMethod is restricted with the one-time generated drafts from the perspective of the imaginary frame and schema items of the preliminary logical forms, which can be further improved with interactively generation and retrieval. Moreover, we have not explored whether the performance can be further improved with explanation/instruction during the stage of draft generation. We will take these limitations into account and mitigate them in future work.

%% file: ethics.tex
% Scientific work published at ACL 2023 must comply with the ACL Ethics Policy.\footnote{\url{https://www.aclweb.org/portal/content/acl-code-ethics}} We encourage all authors to include an explicit ethics statement on the broader impact of the work, or other ethical considerations after the conclusion but before the references. The ethics statement will not count toward the page limit (8 pages for long, 4 pages for short papers).
% \section*{Ethics Statement}